\documentclass[10pt,twocolumn,letterpaper]{article}

\usepackage{cvpr}              

\usepackage{booktabs}
\usepackage{multirow}
\usepackage{makecell}  
\usepackage{threeparttable}
\usepackage{graphicx}  
\usepackage{amssymb} 
\usepackage{graphicx} 
\definecolor{LightBlue}{rgb}{0.4, 0.6, 0.9} 
\definecolor{DarkYellow}{rgb}{0.85, 0.65, 0.0}

\PassOptionsToPackage{dvipsnames,table}{xcolor}
\usepackage{textcomp}
\newcommand{\midtilde}{\raisebox{0.5ex}{\texttildelow}}

\definecolor{cvprblue}{rgb}{0.21,0.49,0.74}
\usepackage[pagebackref,breaklinks,colorlinks,allcolors=cvprblue]{hyperref}

\def\confName{CVPR}
\def\confYear{2025}

\title{
\raisebox{-0.4em}{\includegraphics[width=0.8cm]{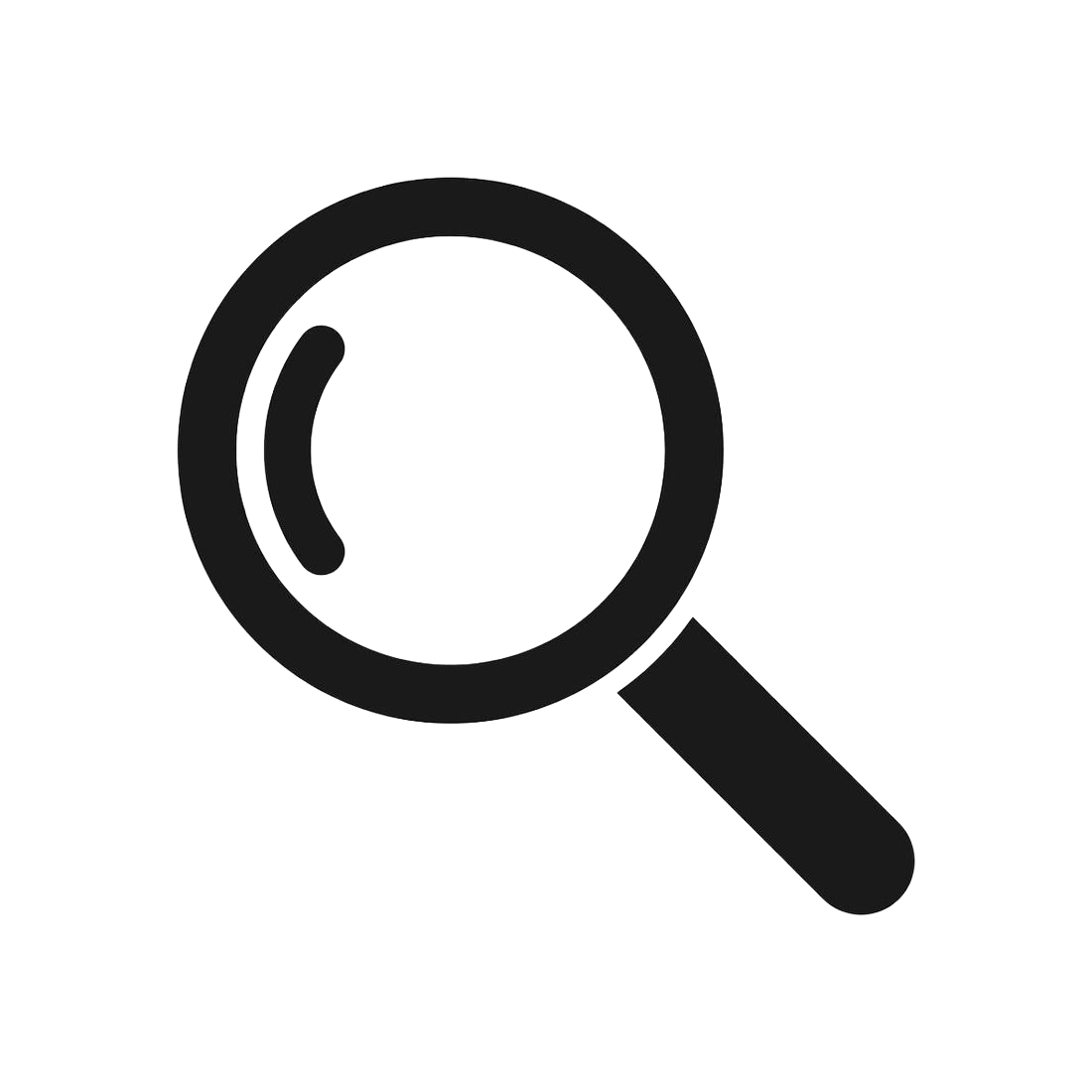}}
\hspace{-0.7em}
\raisebox{-0.8em}{\includegraphics[width=15cm]{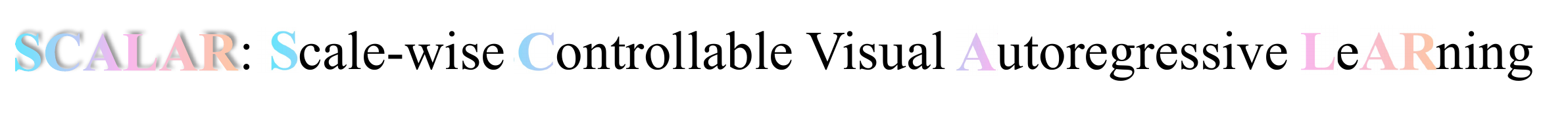}}
\vspace{-2em} 
}

\author{
Ryan Xu$^{*}$,\space\space
Dongyang Jin\footnotemark[1]\space\space, \space
Yancheng Bai\footnotemark[2]\hspace{0.4em}\footnotemark[3]\space,  
Rui Lan,  \\
Xu Duan, \space
Lei Sun\footnotemark[2] \space, 
and Xiangxiang Chu \\
{\normalsize Amap, Alibaba Group}\\
{\tt \small ryansxu.00@gmail.com}, 
{\tt \small \{jindongyang.j, lr264907, xuxu.dx\}@alibaba-inc.com}, \\
{\tt \small  \{yancheng.byc, ally.sl, chuxiangxiang.cxx\}@alibaba-inc.com}
}

\begin{document}

\twocolumn[{%
\renewcommand\twocolumn[1][]{#1}%
\maketitle
\begin{center}
    \centering
    \vspace{-2em} 
    \captionsetup{type=figure}
    \includegraphics[width=2.0\columnwidth]{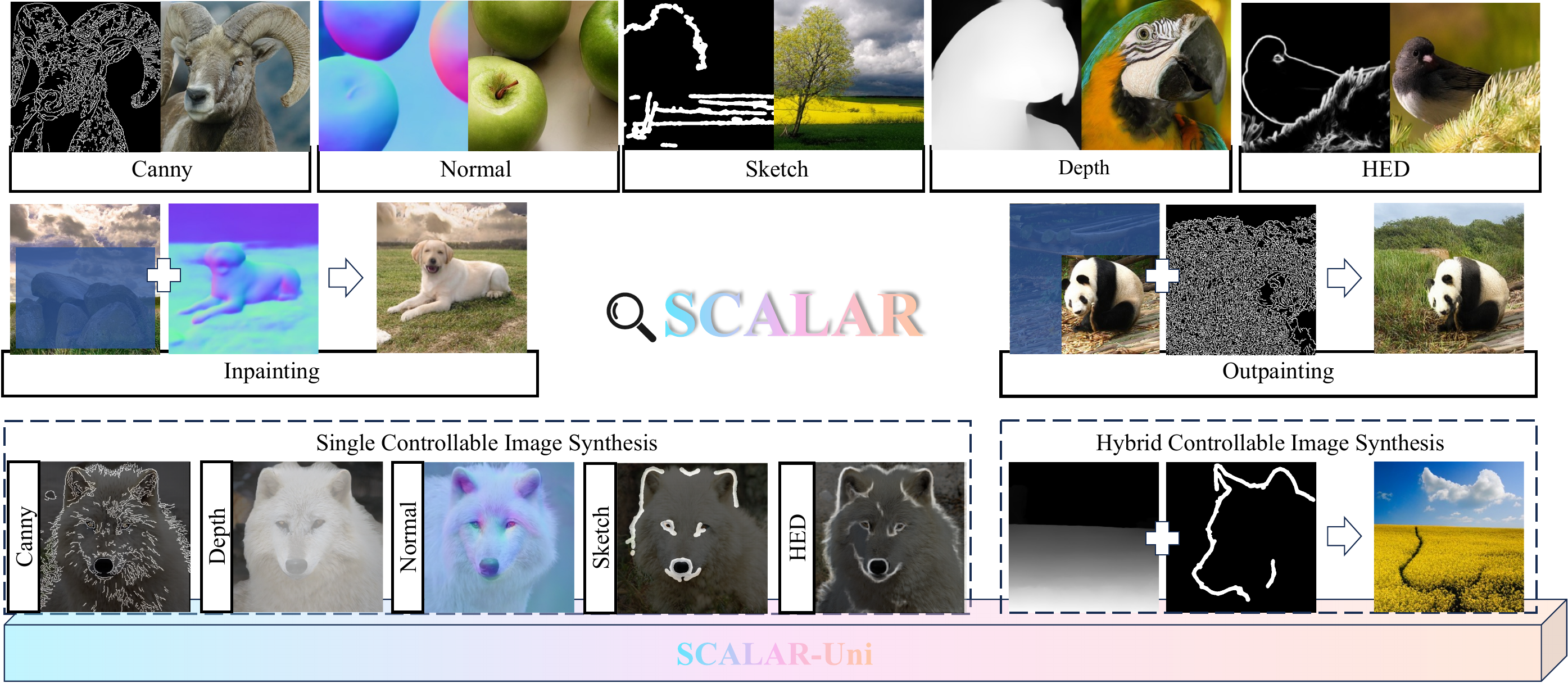}
    \captionof{figure}{
    \textbf{SCALAR}, a novel controllable VAR method, achieves superior generation quality and control capabilities for various types of controllable signals (top row). It also exhibits robust zero-shot generalizability to tasks such as inpainting and outpainting (middle row). \textbf{SCALAR-Uni} further extends it by supporting multi-condition control within a unified model (bottom row).
    }
    \label{fig:f1}
\end{center}%
}]

\footnotetext[1]{Equal contribution.}
\footnotetext[2]{Corresponding author.}
\footnotetext[3]{Project Leader.}


\begin{abstract}
Controllable image synthesis, which enables fine-grained control over generated outputs, has emerged as a key focus in visual generative modeling.
However, controllable generation remains challenging for Visual Autoregressive (VAR) models due to their hierarchical, next-scale prediction style.
Existing VAR-based methods often suffer from inefficient control encoding and disruptive injection mechanisms that compromise both fidelity and efficiency. 
In this work, we present SCALAR, a controllable generation method based on VAR, incorporating a novel Scale-wise Conditional Decoding mechanism. 
SCALAR leverages a pretrained image encoder to extract semantic control signal encodings, which are projected into scale-specific representations and injected into the corresponding layers of the VAR backbone. 
This design provides persistent and structurally aligned guidance throughout the generation process.
Building on SCALAR, we develop SCALAR-Uni, a unified extension that aligns multiple control modalities into a shared latent space, supporting flexible multi-conditional guidance in a single model.
quality and control precision, including diffusion-based methods (e.g., ControlNet++) and AR-based methods (e.g., ControlAR, CAR).
The code is released at \href{https://github.com/AMAP-ML/SCALAR}{https://github.com/AMAP-ML/SCALAR}.
\end{abstract}

\section{Introduction}
\label{sec:intro}
Controllable image synthesis is a pivotal domain in visual generation, enabling the precise and nuanced creation of visual content according to specific user guidance. Recent advances in this field are currently dominated by two primary paradigms: Diffusion Models~\cite{ddpm,song2020score} and Autoregressive Models (AR)~\cite{llamagen,var}. While diffusion-based methods~\cite{controlnet,controlnet++,unicontrol} have achieved widespread success, the iterative denoising process is not inherently compatible with the sequential, token-based architecture of LLMs, which hinders the development of truly unified multimodal modeling~\cite{zhao2023easygen,wang2024multi}.

Visual Autoregressive (VAR) models~\cite{var}, a leading approach within the autoregressive paradigm, offer a compelling path forward. By framing image synthesis as a next-scale prediction task, VAR-based models~\cite{han2025infinity,tang2024hart,ma2024star,zhuang2025vargpt} align naturally with LLM architectures and have demonstrated better inference efficiency and generative quality than both state-of-the-art diffusion models~\cite{sdxl} and raster-scan-based autoregressive models~\cite{llamagen}. 
Nevertheless, the capacity of VAR models for fine-grained control remains a significant and underexplored challenge. This challenge primarily stems from the unique hierarchical, scale-wise generation process, which presents a distinct paradigm from existing approaches: In diffusion models, control signals are applied globally to the denoising network at each step, whereas in traditional raster-scan AR models~\cite{llamagen, aim}, they are injected before each spatial token prediction. 

Existing works on controllable Visual Autoregressive learning, notably CAR~\cite{yao2024car} and ControlVAR~\cite{controlvar}, deliver suboptimal performance. We attribute this bottleneck to two fundamental design flaws. First, they utilize complex and disruptive injection mechanisms; architectures with parallel branches~\cite{yao2024car} similar to  ControlNet or joint control and image modeling schemes~\cite{controlvar} not only incur substantial computational overhead but also implicitly disrupt the powerful generative capabilities of the pretrained backbone.
Second, they often utilize lightweight convolutional networks or VQ-VAEs as control encoders, which have been proven to have a limited capacity in capturing rich spatial-semantic features~\cite{zhou2025dino,li2024imagefolder}. Consequently, their performance in both generation quality and control consistency trails behind even raster-scan AR models~\cite{li2024controlar}, highlighting a critical need for a control mechanism that is both efficient and fundamentally aligned with the native structure of VAR.

To this end, we present SCALAR, an effective controllable generation method based on VAR. Aligning with the more efficient paradigm~\cite{li2024controlar}, SCALAR injects control during the autoregressive decoding phase. Our approach is centered on a novel and simple \textbf{Scale-wise Conditional Decoding} mechanism. To be specific, we first employ a pretrained vision foundation model~\cite{dinov2, sam, vit} to extract powerful, scale-agnostic control features that contain rich semantic information. These general features are then processed by scale-wise lightweight projection blocks—which maintain independent weights for each scale—to produce specialized \textbf{Control Signal Encodings}. Ultimately, SCALAR injects these tailored encodings directly into the hidden states of the VAR backbone’s layers, providing scale-wise persistent guidance throughout the next-scale prediction process. 
Building on our design, we extend SCALAR to a unified control version, SCALAR-uni. This extension incorporates a \textbf{Unified Control Alignment} process that projects features from diverse control modalities into a common latent space, enabling seamless guidance of various condition types from a unified model.

The main contributions of this paper are as follows:  
\begin{itemize}
    \item We present SCALAR, a controllable generation method based on VAR that introduces a Scale-wise Conditional Decoding mechanism, explicitly designed to align with the next-scale prediction nature of VAR models. 
    \item A Unified Control Alignment process is introduced to SCALAR-Uni, enabling the various condition semantics guidance in controllable visual autoregressive learning.
    \item Extensive experiments demonstrate that our proposed SCALAR exhibits exceptional generation quality and conditional consistency across both class-to-image (c2i) and text-to-image (t2i) controllable generation, significantly outperforming other SoTA methods.
\end{itemize}

\section{Related Work}
\subsection{Image Generation}
Among recent advances in image generation, two paradigms have gained particular traction: diffusion models and autoregressive (AR) models.
Diffusion-based methods, beginning with DDPM~\cite{ho2020denoising}, approach generation by progressively denoising Gaussian noise.
This framework has since evolved significantly, with improvements in sampling techniques and latent representations~\cite{nichol2021improved, rombach2022high, zhang2025robust, peng2024efficient, peng2025pixel} boosting both visual quality and efficiency.
Such models now power the majority of high-fidelity text-to-image and text-to-video systems~\cite{saharia2022photorealistic, singer2022make, ma2024passersby, lan2025flux, peng2025boosting, zhang2025boow, ragsr, peng2024unveiling}, typically adopting a U-Net architecture for denoising and using CLIP or T5 as the backbone for text understanding.
More recently, diffusion transformers like DiT~\cite{dit} have demonstrated the scalability of replacing U-Nets with Transformer backbones.
Despite their success, diffusion models are known for high computational costs, prompting a shift of attention toward AR models as a more efficient alternative.

\begin{figure*}[!t]
\centering
\includegraphics[width=2.0\columnwidth]{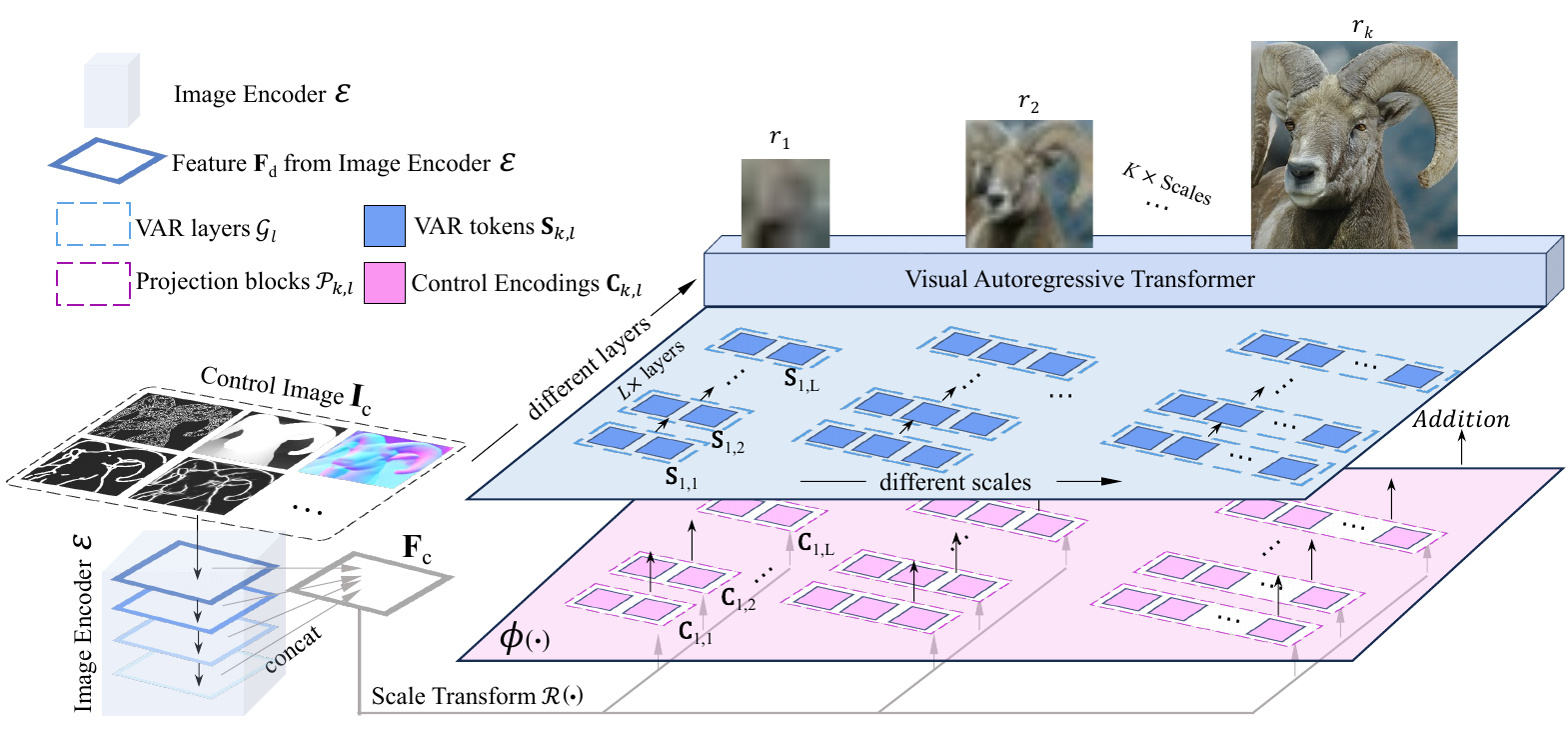}
\caption{
The framework of our \textbf{SCALAR} applies a next-scale paradigm adapted for VAR to design a Scale-wise Conditional Decoding mechanism (see~\Cref{sec:scale_wise_conditional_decoding} for details). The feature $\mathbf{F}_{c}$ is obtained by concatenating four features $\mathbf{F}_{d}$ extracted by the Image Encoder $\mathcal{E}$.
}
\label{fig:pipeline}
\end{figure*}

Autoregressive (AR) models take a different route, casting image generation as a sequential prediction task.
Initial methods~\cite{van2016pixel} directly modeled raw pixel outputs in raster-scan order.
Recent developments, inspired by the progress of large language models~\cite{touvron2023llama, achiam2023gpt}, introduce discrete visual representations using tokenizers like VQ-VAE~\cite{van2017neural} and VQ-GAN~\cite{esser2021taming}, allowing AR models to operate on compact image tokens and predict them autoregressively.
Recent works~\cite{an2025unictokens,an2024mc,luo2024llm,lin2025perceive,lin2024draw} such as LlamaGen~\cite{llamagen} and Open-MAGVIT2~\cite{open-magvit2} adopt the LLaMA architecture~\cite{llama}, while AiM~\cite{aim} explores the Mamba~\cite{mamba} as the autoregressive backbone.
Among them, Visual Autoregressive Modeling (VAR) introduces a scalable next-scale prediction mechanism, which differs from traditional raster-scan AR methods.
VAR-based models for class-to-image and text-to-image tasks~\cite{var, ma2024star, tang2024hart, han2025infinity, zhuang2025vargpt} achieve image synthesis performance comparable to state-of-the-art diffusion models while offering significant computational efficiency.

\subsection{Controllable Image Generation}
Conditional image generation, where models are guided by external signals, has become a central topic in generative modeling.
Early efforts focused on class labels and attributes, explored via GANs~\cite{mirza2014conditional,song2023fashion} and VAEs~\cite{kingma2013auto,zeng2024cat}. More recent diffusion-based methods~\cite{controlnet,t2i-adapter} integrate control signals through cross-attention or adapter modules, enabling high-quality generation without full-model fine-tuning. Follow-up work~\cite{unicontrol,controlnet++,ran2024x} further improves generalization across control types and model architectures.

In contrast, conditional generation for AR models has been far less explored. Existing works on controllable autoregressive learning are typically classified according to the guidance injection phase in autoregressive generation: pre-filling phase or decoding phase. 
Pre-filling-phase methods \cite{controlvar, qu2025varsr} fill the control encodings into the initial sequence from the start. 
This approach increases the token sequence length, inflating the computational load, while the joint modeling scheme in ControlVAR \cite{controlvar} can disrupt the powerful capabilities of the pretrained backbone. 
Decoding-phase methods \cite{li2024controlar}, which continuously inject control encodings throughout the step-by-step decoding, are simpler and more flexible. However, CAR \cite{yao2024car} adopts parallel conditional decoding branches inspired by ControlNet, which introduces huge complexities when autoregressive decoding. Notably, although VAR models generally outperform traditional AR models in image synthesis quality, these VAR-based controllable methods~\cite{controlvar,yao2024car} still underperform compared to raster-scan AR-based methods~\cite{li2024controlar} in terms of control consistency and image quality. 
This gap indicates that the potential of VAR in controllable generation is yet to be fully explored. 
In this paper, we aim to develop a general and efficient method for controllable image generation based on VAR.

\section{SCALAR}
\label{sec:sc3}
In this section, we present SCALAR, a simple and efficient method for controllable VAR. We first revisit the image generation with VAR. Next, we introduce Scale-wise Conditional Decoding. Then, we conduct several experiments to find the model architecture that yields the best generation quality and control consistency. Finally, we extend our approach to a unified version, SCALAR-Uni, with the proposed Unified Control Alignment.

\begin{figure*}[!t]
\centering
\includegraphics[width=2.0\columnwidth]{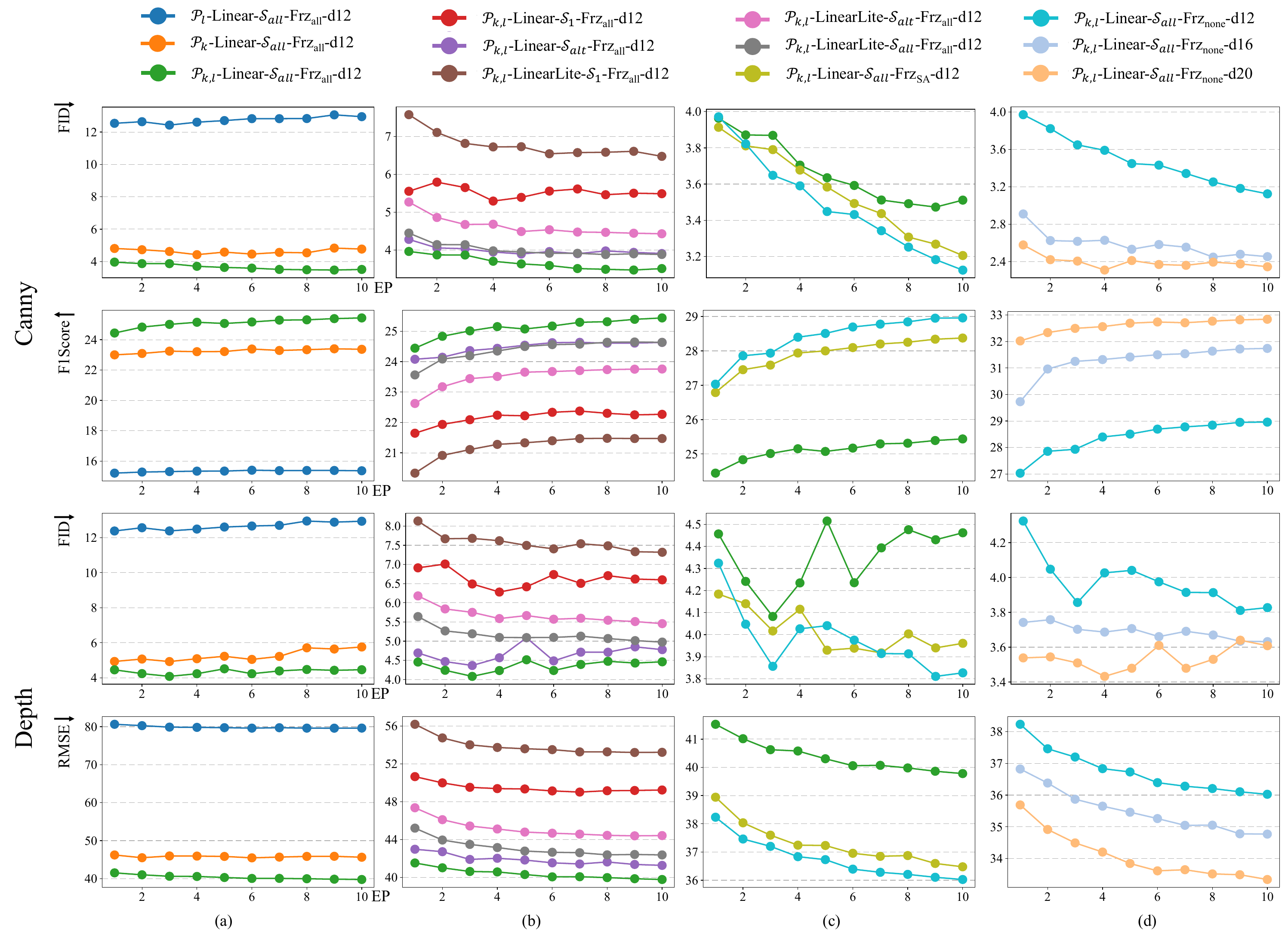}
\caption{
(a) Comparison of parameter sharing for projection blocks ($\mathcal{P}_{k,l}$, $\mathcal{P}_{k}$, and $\mathcal{P}_{l}$).
(b) Comparison of various injection layers set ($\mathcal{S}_1$, $\mathcal{S}_{\text{alt}}$, and $\mathcal{S}_{\text{all}}$) with different structures of projection block (\textbf{Linear} and \textbf{LinearLite}). 
(c) Comparison of different parameter-efficient training strategies ($\text{Frz}_\text{none}$, $\text{Frz}_\text{SA}$, and $\text{Frz}_\text{all}$).
(d) Impacts of scaling up the depth of VAR backbone (VAR-d12, d16, and d20).
\textbf{Note:} All experiments are conducted on ImageNet~\cite{imagenet} with the c2i controllable generation.
}
\label{fig:linechart}
\end{figure*}

\subsection{Preliminary}
Building upon the success of large language models~\cite{touvron2023llama,achiam2023gpt}, autoregressive models adopt discrete quantizers such as VQVAE to convert image patches into index-wise tokens, enabling image generation via next-token prediction over visual token sequences. 
Among them, VAR introduces a scalable next-scale prediction mechanism to generate images across resolutions.
VAR-based class-to-image (c2i)~\cite{var} and text-to-image (t2i)~\cite{ma2024star,tang2024hart,han2025infinity,zhuang2025vargpt} models achieve image synthesis performance comparable to state-of-the-art diffusion models, while reducing computational cost.

The core idea of the VAR model lies in next-scale prediction, which contrasts with traditional raster-scan AR models based on next-token prediction. While conventional AR models generate images token by token along a flattened pixel sequence, this formulation suffers from mathematical inconsistencies and often leads to structural degradation, especially in highly structured images.
In contrast, VAR avoids these issues by operating at the level of token maps, predicting the image progressively across multiple scales. Starting from a coarse $1 \times 1$ token map $r_1$, it autoregressively generates a sequence of higher-resolution token maps $(r_2, \dots, r_K)$, each representing a finer level of detail. The overall generation process is formulated as:
\begin{align}
p(r_1, r_2, \dots, r_K) = \prod_{k=1}^{K} p(r_k \mid r_{<k}),
\end{align}
where $r_k \in [V]^{h_k \times w_k}$ represents the token map at scale $k$, with dimensions $h_k$ and $w_k$, conditioned on previous maps $r_{<k}$. Each token in $r_k$ is an index from the VQVAE codebook $V$, which is trained through multi-scale quantization and shared across scales.
A standard cross-entropy loss is used to supervise VAR, defined as:
\begin{equation}
\mathcal{L}_{CE} = \mathbb{E}_{r_k \sim p_(r_k)} \left[ - \log p_{\theta}(r_k \mid r_{<k}) \right].
\end{equation}
The loss is applied at each scale to train the model to predict finer token maps conditioned on coarser ones.

\subsection{Scale-wise Conditional Decoding}
\label{sec:scale_wise_conditional_decoding}
The generation process in autoregressive models is bifurcated into two phases: pre-filling and decoding. In SCALAR, we diverge from the conditional pre-filling way~\cite{controlvar, qu2025varsr}, which relies on a distinct initial conditioning filling step. 
We adopt a simpler conditional decoding strategy~\cite{li2024controlar, yao2024car}. This design integrates the control signal directly and continuously throughout the decoding process. The scale-autoregressive nature of VARs poses distinct requirements for how control signals are applied across the generative hierarchy; guidance must be present and adapted at each scale. SCALAR achieves this by injecting the control signal into a predefined subset of layers, indexed by $l \in \mathcal{S}$. Formally, the operation at each layer is expressed as:
\begin{equation}
\mathbf{S}_{k,l}^{\prime} = \mathcal{G}_l(\mathbf{S}_{k,l}+\mathbf{C}_{k,l}),
\end{equation}
where $\mathcal{G}_{l}$ represents the $l$-th layer of the GPT-style decoder transformer, $\mathbf{S}_{k,l}$ and $\mathbf{S}_{k,l}^{\prime}$ denote the input and output sequence of image tokens at layer $l$ for scale $k$, and $\mathbf{C}_{k,l}$ is the corresponding control signal encoding sequence injected at that same layer and scale. The set $\mathcal{S} \subseteq \{0, \dots, L-1\}$ contains the indices of the layers targeted for injection, where $L$ is the total depth of the decoder $\mathcal{G}$.

\noindent\textbf{Control Signal Encoding $\mathbf{C}_{k,l}$.} We employ a pretrained vision foundation model~\cite{dinov2, sam, vit} as our universal control signal encoder. Its inherent understanding of rich visual semantics, learned via large-scale self-supervision, provides a powerful and robust feature extractor for diverse control conditions.
To leverage the hierarchical features within the encoder~\cite{bolya2025meta_perception_encoder}, the control representation $\mathbf{F}_c$ is formed by concatenating features from the $i$-th layers of the encoder $\mathcal{E}$. It can be expressed as:
\begin{equation}
\mathbf{F}_c = {\mathcal{C}} \left( \underset{i \in \mathcal{I}}{\mathcal{E}_i}(\mathbf{I}_{c}) \right),
\end{equation}
where $\mathcal{E}_i(\cdot)$ is the feature from the $i$-th layer of the encoder, $\mathcal{C}(\cdot)$ represents the concatenation operation and $\mathcal{I}$ is the set of indices for the selected layers. The resulting scale-agnostic feature $\mathbf{F}_c$ is then tailored for injection into scale $k$ and layer $l$ of our Scale-wise Conditional Decoding process:
\begin{align}
\mathbf{C}_{k,l} = \phi_{k,l}(\mathbf{F}_c) = \mathcal{P}_{k,l}(\mathcal{R}_{h_k,w_k}(\mathbf{F}_c)),
\end{align}
where $\mathcal{R}_{h_k,w_k}$ denotes transforming the features to the target spatial size $(h_k,w_k)$ of scale $k$. Subsequently, a conditional projection block $\mathcal{P}_{k,l}$ maps to the final control signal encoding $\mathbf{C}_{k,l}$.

\begin{table*}[!t]
\centering
\caption{
Quantitative results of conditional image generation on ImageNet~\cite{deng2009imagenet}. 
Values marked with \midtilde\ are estimated from histograms in the corresponding paper.
Cells highlighted with \colorbox[HTML]{E6F2FF}{\phantom{xx}} indicate the best performance for each setting.
}
\vspace{-0.4em}
\resizebox{2.0\columnwidth}{!}{ 
\setlength{\tabcolsep}{1mm}
\begin{tabular}{c|c|c|ccc|ccc|ccc|ccc|ccc}
\toprule
\multirow{2}{*}{Type}  & \multirow{2}{*}{Method}                         & \multirow{2}{*}{Model} & \multicolumn{3}{c|}{Canny} & \multicolumn{3}{c|}{Depth} & \multicolumn{3}{c|}{Normal} & \multicolumn{3}{c|}{HED} & \multicolumn{3}{c}{Sketch} \\
                       &                                                 &                        & FID↓  & IS↑   & F1-Score↑  & FID↓  & IS↑   & RMSE↓      & FID↓  & IS↑   & RMSE↓       & FID↓ & IS↑    & SSIM↑    & FID↓  & IS↑ & F1-Score↑ \\
\midrule
\multirow{2}{*}{Diff.} & T2IAdapter~\cite{mou2024t2i}                    & -                      & \midtilde 10.2 & \midtilde 157 & -          & \midtilde 9.9  & \midtilde 134 & -          & \midtilde 9.5  & \midtilde 143 & -           & \midtilde 9.3  & \midtilde 142 & -        & \midtilde 16.2 & \midtilde 156 & - \\
                       & ControlNet~\cite{zhang2023adding}               & -                      & \midtilde 11.6 & \midtilde 173 & -          & \midtilde 9.2  & \midtilde 150 & -          & \midtilde 8.9  & \midtilde 155 & -           & \midtilde 8.6  & \midtilde 150 & -        & \midtilde 15.3 & \midtilde 163 & - \\
\midrule

\multirow{3}{*}{AR}    & \multirow{3}{*}{ControlAR~\cite{li2024controlar}} & AiM-L                & 9.66  & -     & 30.36      & 7.39  & -     & 35.01      &       & -     &             &       & -     &          &       & -     &   \\
                       &                                                 & LlamaGen-B             & 10.64 & -     & 34.15      & 6.67  & -     & 32.41      &       & -     &             &       & -     &          &       & -     &   \\
                       &                                                 & \cellcolor[rgb]{0.9,0.95,1}LlamaGen-L             & \cellcolor[rgb]{0.9,0.95,1}7.69  & \cellcolor[rgb]{0.9,0.95,1}-     & \cellcolor[rgb]{0.9,0.95,1}34.91      & \cellcolor[rgb]{0.9,0.95,1}4.19  & \cellcolor[rgb]{0.9,0.95,1}-     & \cellcolor[rgb]{0.9,0.95,1}31.11      & \cellcolor[rgb]{0.9,0.95,1}~      & \cellcolor[rgb]{0.9,0.95,1}-     & \cellcolor[rgb]{0.9,0.95,1}~            & \cellcolor[rgb]{0.9,0.95,1}~      & \cellcolor[rgb]{0.9,0.95,1}-     & \cellcolor[rgb]{0.9,0.95,1}~         & \cellcolor[rgb]{0.9,0.95,1}~      & \cellcolor[rgb]{0.9,0.95,1}-     & \cellcolor[rgb]{0.9,0.95,1}~  \\
\midrule
\multirow{13}{*}{VAR}   & \multirow{5}{*}{ControlVAR~\cite{controlvar}}   & VAR-d12                & \midtilde 35.2 & \midtilde 44  & -          & \midtilde 26.7 & 52  & -          & \midtilde 25.3 & \midtilde 55  & -           &       & -     &          &       & -     &   \\
                       &                                                 & VAR-d16                & \midtilde 16.2 & \midtilde 80  & -          & \midtilde 13.8 & 92  & -          & \midtilde 14.2 & \midtilde 89  & -           &       & -     &          &       & -     &   \\
                       &                                                 & VAR-d20                & \midtilde 13.0 & \midtilde 94  & -          & \midtilde 13.4 & 98  & -          & \midtilde 12.8 & \midtilde 100 & -           &       & -     &          &       & -     &   \\
                       &                                                 & VAR-d24                & \midtilde 15.7 & \midtilde 100 & -          & \midtilde 12.5 & 125 & -          & \midtilde 11.8 & \midtilde 123 & -           &       & -     &          &       & -     &   \\
                       &                                                 & \cellcolor[rgb]{0.9,0.95,1}VAR-d30                & \cellcolor[rgb]{0.9,0.95,1}7.85  & \cellcolor[rgb]{0.9,0.95,1}160.0 & \cellcolor[rgb]{0.9,0.95,1}-          & \cellcolor[rgb]{0.9,0.95,1}6.50  & \cellcolor[rgb]{0.9,0.95,1}180.5 & \cellcolor[rgb]{0.9,0.95,1}-          & \cellcolor[rgb]{0.9,0.95,1}6.20  & \cellcolor[rgb]{0.9,0.95,1}172.0 & \cellcolor[rgb]{0.9,0.95,1}-           & \cellcolor[rgb]{0.9,0.95,1}~      & \cellcolor[rgb]{0.9,0.95,1}-     & \cellcolor[rgb]{0.9,0.95,1}~         & \cellcolor[rgb]{0.9,0.95,1}~      & \cellcolor[rgb]{0.9,0.95,1}-     & \cellcolor[rgb]{0.9,0.95,1}~  \\
                       \cline{2-18}
                       & \multirow{4}{*}{CAR~\cite{yao2024car}}          & VAR-d16                & \midtilde 12.8 & \midtilde 85  & -          & \midtilde 10.8 & \midtilde 95  & -          & \midtilde 11.0 & \midtilde 98  & -           & \midtilde 9.8  & \midtilde 102 & -        & \midtilde 13.2 & \midtilde 83  & - \\
                       &                                                 & VAR-d20                & \midtilde 10.2 & \midtilde 125 & -          & \midtilde 8.0  & \midtilde 135 & -          & \midtilde 8.8  & \midtilde 138 & -           & \midtilde 7.2  & \midtilde 144 & -        & \midtilde 11.2 & \midtilde 118 & - \\
                       &                                                 & VAR-d24                & \midtilde 9.0  & \midtilde 155 & -          & \midtilde 7.0  & \midtilde 165 & -          & \midtilde 7.5  & \midtilde 168 & -           & \midtilde 6.5  & \midtilde 176 & -        & \midtilde 10.0 & \midtilde 143 & - \\
                       &                                                 & \cellcolor[rgb]{0.9,0.95,1}VAR-d30                & \cellcolor[rgb]{0.9,0.95,1}8.30  & \cellcolor[rgb]{0.9,0.95,1}167.3 & \cellcolor[rgb]{0.9,0.95,1}-          & \cellcolor[rgb]{0.9,0.95,1}6.90  & \cellcolor[rgb]{0.9,0.95,1}178.6 & \cellcolor[rgb]{0.9,0.95,1}-          & \cellcolor[rgb]{0.9,0.95,1}6.60  & \cellcolor[rgb]{0.9,0.95,1}175.9 & \cellcolor[rgb]{0.9,0.95,1}-           & \cellcolor[rgb]{0.9,0.95,1}5.60  & \cellcolor[rgb]{0.9,0.95,1}182.2 & \cellcolor[rgb]{0.9,0.95,1}-        & \cellcolor[rgb]{0.9,0.95,1}10.20 & \cellcolor[rgb]{0.9,0.95,1}161.6 & \cellcolor[rgb]{0.9,0.95,1}- \\
                       \cline{2-18}
                       & \multirow{4}{*}{\shortstack{SCALAR\\(Ours)}}     & VAR-d12               & 3.12  & 191.3 &  28.96     & 3.83  & 233.3  & 36.03      & 3.76  & 225.7 &  28.14      & 2.62  & 189.1 & 74.93    & 4.55  & 229.6 & 76.60    \\
                       &                                                 & VAR-d16                & 2.45  & 237.7 &  31.74     & 3.63  & 285.8  & 34.77      & 3.70  & 279.5 &  27.64      & 1.97  & 222.5 & 76.37    & 4.51  & 292.4 & 77.03    \\
                       &                                                 & VAR-d20                & 2.34  & 254.3 &  32.84     & 3.61  & 301.3  & 33.34      & 3.51  & 300.9 &  27.57      & 1.81  & 240.4 & 76.74    & 4.22  & 312.7 & 77.43    \\
                       &                                                 &\cellcolor[rgb]{0.9,0.95,1} VAR-d24                & \cellcolor[rgb]{0.9,0.95,1}2.14 & \cellcolor[rgb]{0.9,0.95,1}261.8 &  \cellcolor[rgb]{0.9,0.95,1}33.14     &\cellcolor[rgb]{0.9,0.95,1}3.09   & \cellcolor[rgb]{0.9,0.95,1}306.9 & \cellcolor[rgb]{0.9,0.95,1}33.13 & \cellcolor[rgb]{0.9,0.95,1}3.09 & \cellcolor[rgb]{0.9,0.95,1}307.2  & \cellcolor[rgb]{0.9,0.95,1}27.45       & \cellcolor[rgb]{0.9,0.95,1}1.72   & \cellcolor[rgb]{0.9,0.95,1}244.2 & \cellcolor[rgb]{0.9,0.95,1}76.98    & \cellcolor[rgb]{0.9,0.95,1}3.57  & \cellcolor[rgb]{0.9,0.95,1}315.5 & \cellcolor[rgb]{0.9,0.95,1}77.58    \\
                       \cline{2-18}
                       & \multirow{3}{*}{\shortstack{SCALAR-Uni\\(Ours)}}& VAR-d12               & 3.31 & 207.4 & 27.52 & 4.21  & 231.5  & 38.57 & 4.05  & 222.5  & 29.74  & 2.82  & 218.4 & 71.10 & 4.75  & 222.9 & 75.40    \\ 
                       &                                                 & VAR-d16               & 2.78 & 262.4 & 30.76 & 3.73  & 294.1  & 37.24 & 4.03  & 288.6  & 29.17  & 2.53  & 263.9 & 72.73 & 4.58  & 297.8 & 75.66   \\
                       &                                                 & \cellcolor[rgb]{0.9,0.95,1}VAR-d20 & \cellcolor[rgb]{0.9,0.95,1}2.64 & \cellcolor[rgb]{0.9,0.95,1}276.9 & \cellcolor[rgb]{0.9,0.95,1}31.68 & \cellcolor[rgb]{0.9,0.95,1}3.69  & \cellcolor[rgb]{0.9,0.95,1}310.1  & \cellcolor[rgb]{0.9,0.95,1}37.12 & \cellcolor[rgb]{0.9,0.95,1}3.56  & \cellcolor[rgb]{0.9,0.95,1}308.5  & \cellcolor[rgb]{0.9,0.95,1}29.48  & \cellcolor[rgb]{0.9,0.95,1}2.37  & \cellcolor[rgb]{0.9,0.95,1}272.6 & \cellcolor[rgb]{0.9,0.95,1}72.82 & \cellcolor[rgb]{0.9,0.95,1}4.05  & \cellcolor[rgb]{0.9,0.95,1}311.8 & \cellcolor[rgb]{0.9,0.95,1}75.76   \\
\bottomrule
\end{tabular}
}
\vspace{-0.2em}
\label{tab:main_result}
\end{table*}

\subsection{Exploring Controlled Architectural Design}

 We explore several strategies to determine the optimal architecture for SCALAR. Our primary goal is to identify the most effective designs for encoding and injecting control signals into the VAR's generative process. We focus our experiments on the following key architectural observations:

\begin{itemize}
    \item \textbf{Parameter Sharing of $\mathcal{P}_{k,l}$}. The control signal is injected via projection blocks. We investigate different parameter-sharing strategies for these layers. Specifically, we explore whether the projection weights should be: (i) unique for each transformer layer at each scale, marked as $\mathcal{P}_{k,l}$, (ii) shared across all layers, marked as $\mathcal{P}_{k}$, or (iii) shared across all scales, marked as $\mathcal{P}_{l}$.
    \item \textbf{Projection Blocks $\mathcal{P}_{k,l}$ and Injection Layers Set $\mathcal{S}$.} By default, $\mathcal{P}_{k,l}$ is a single linear layer, which can be parameter-intensive, especially when the injection set $\mathcal{S}$ is large. We explore two main aspects:
    \begin{itemize}
        \item \textbf{Structure of $\mathcal{P}_{k,l}$}: We compare (i) \textbf{Linear}: a standard linear layer against (ii) \textbf{LinearLite}: a more parameter-efficient bottleneck structure, which consists of two linear layers that first squeeze the channel dimension and then expand it back.
        \item \textbf{Injection Set $\mathcal{S}$}: We investigate how the density of control signal injection affects performance by testing three configurations for $\mathcal{S}$: (i) injecting only into the first layer $\mathcal{S}_1=\{1\}$, (ii) injecting into alternating layers $\mathcal{S}_{\text{alt}}=\{1, 3, 5, \dots, L-1\}$, and (iii) injecting into all layers $\mathcal{S}_{\text{all}}=\{1, 2, 3, \dots, L\}$.
    \end{itemize}

    \item \textbf{Parameter-Efficient Training}. We analyze various training strategies to evaluate the trade-off between performance and the number of trainable parameters. Specifically, we compare: (i) $\text{Frz}_\text{none}$: fine-tuning all VAR parameters, (ii) $\text{Frz}_\text{SA}$: freeze all self-attention layers of VAR, and (iii) $\text{Frz}_\text{all}$: freeze all VAR parameters.
\end{itemize}

For all comparisons above, we train VAR-d12 models on ImageNet $256\times256$ with batch size 512. 
We train models with the above changes and compare them on Fréchet Inception Distance (FID), Inception Score (IS), Root Mean Square Error (RMSE, conditioned on Depth), and F1-Score (conditioned on Canny).
As shown in \Cref{fig:linechart}(a) and (b), the optimal configuration employs projection blocks with scale-wise parameters for each block ($\mathcal{P}_{k,l}$), uses a standard linear layer (\textbf{Linear}), and injects the control signal into all layers ($\mathcal{S}_{\text{all}}$). \Cref{fig:linechart}(c) shows that freezing backbone parameters causes a significant reduction of control performance. In \Cref{fig:linechart}(d), scaling up depth of VAR helps performance. 
Accordingly, we adopt the best combination of settings ($\mathcal{P}_{k,l}$-\textbf{Linear}-$\mathcal{S}_{\text{all}}$-$\text{Frz}_\text{none}$) for our final SCALAR model to achieve the best performance.
Further analysis details about~\Cref{fig:linechart} are provided in the \textbf{Supplementary Material}.

\begin{figure}[!t]
\centering
\includegraphics[width=1.0\columnwidth]{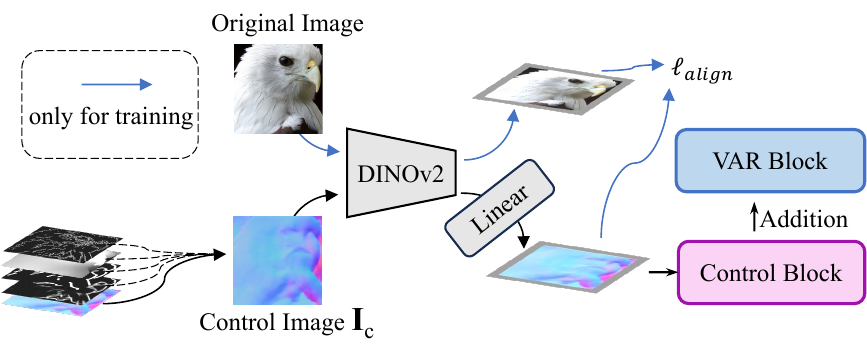}
\caption{
Our SCALAR-Uni, a unified multi-condition control method. Building on SCALAR, we introduce unified control alignment to map diverse control features into a common, modality-agnostic latent space. During training, control images are randomly sampled with equal probability.
}
\label{fig:scalar-uni}
\end{figure}

\subsection{Unified Control Alignment}
\label{sec:unified_control}

Benefiting from the architectural simplicity of SCALAR, extending it to manage multiple conditions simultaneously only requires addressing a key challenge: the control features extracted for different modalities (e.g., Canny, Depth) reside in distinct and potentially incompatible feature spaces.
As shown in \Cref{fig:scalar-uni}, we propose a unified version, \textbf{SCALAR-Uni}. The core idea is to map disparate control features into a common, modality-agnostic latent space. We utilize the image feature space itself as the target for this alignment, as it offers a rich and universal representation of visual concepts. We enforce this alignment by introducing an auxiliary loss term during training, which minimizes the L2 distance between the projected control representations and the corresponding image features:
\begin{align}
    \mathcal{L}_{\text{align}} &= \left\| \mathcal{F}_{\text{align}}(\mathbf{F}_{c}) - \mathbf{F}_{\text{img}} \right\|_{2}^{2}  \nonumber \\
    &= \left\| \mathcal{F}_{\text{align}}(\mathbf{F}_{c}) - {\mathcal{C}} \left( \underset{i \in \mathcal{I}}{\mathcal{E}_i}(\mathbf{I}_{\text{img}}) \right) \right\|_{2}^{2}
    \label{eq:loss_align}
\end{align}
where $\mathbf{I}_{\text{img}}$ represents the corresponding image for control signal, $\mathcal{F}_{\text{align}}$ represents a lightweight alignment module, implemented as a linear layer, learning to project the initial control representation $\mathbf{F}_c$ into this shared image feature space. The total training loss of SCALAR-Uni can be formulated as:
\begin{align}
    \mathcal{L} = \mathcal{L}_{\text{CE}} + \lambda \mathcal{L}_{\text{align}},
\end{align}
where $\lambda$ is a constant that regulates the alignment loss.

\begin{figure*}[!t]
\centering
\includegraphics[width=2.0\columnwidth]{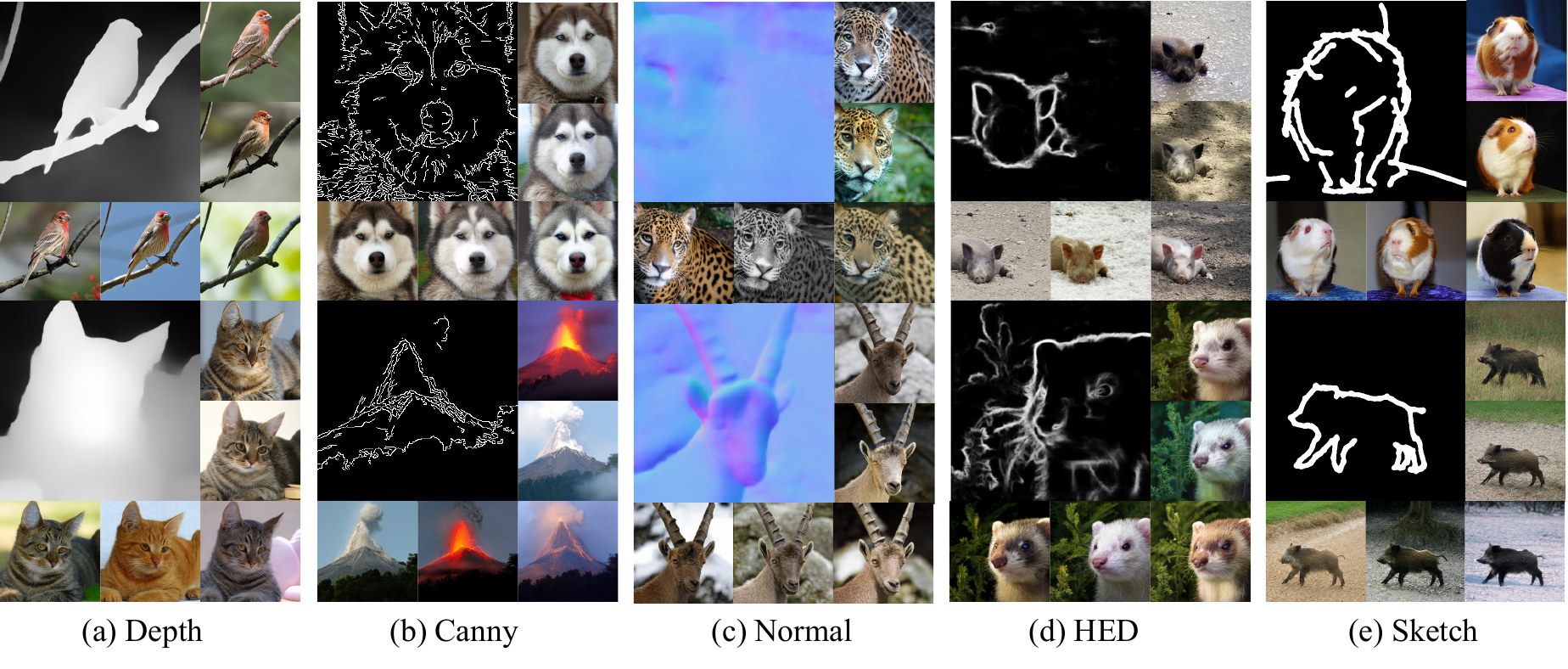}
\vspace{-1.5em}
\caption{
Visual results generated by SCALAR for class-to-image controllable generation.
}
\vspace{-1.2em}
\label{fig:scalar_vis}
\end{figure*}

\begin{figure}[!t]
\centering
\includegraphics[width=1.0\columnwidth]{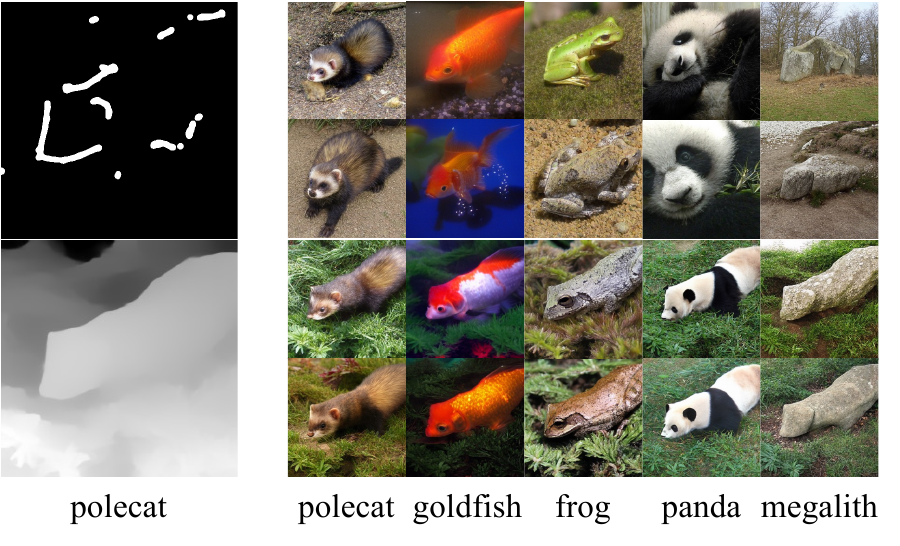}
\vspace{-1.8em}
\caption{
Visualization of class-to-image controllable generation with SCALAR by changing class labels (e.g., “polecat” → “goldfish”).
}
\label{fig:cls}
\end{figure}

\begin{figure}[!t]
\centering
\includegraphics[width=1.0\columnwidth]{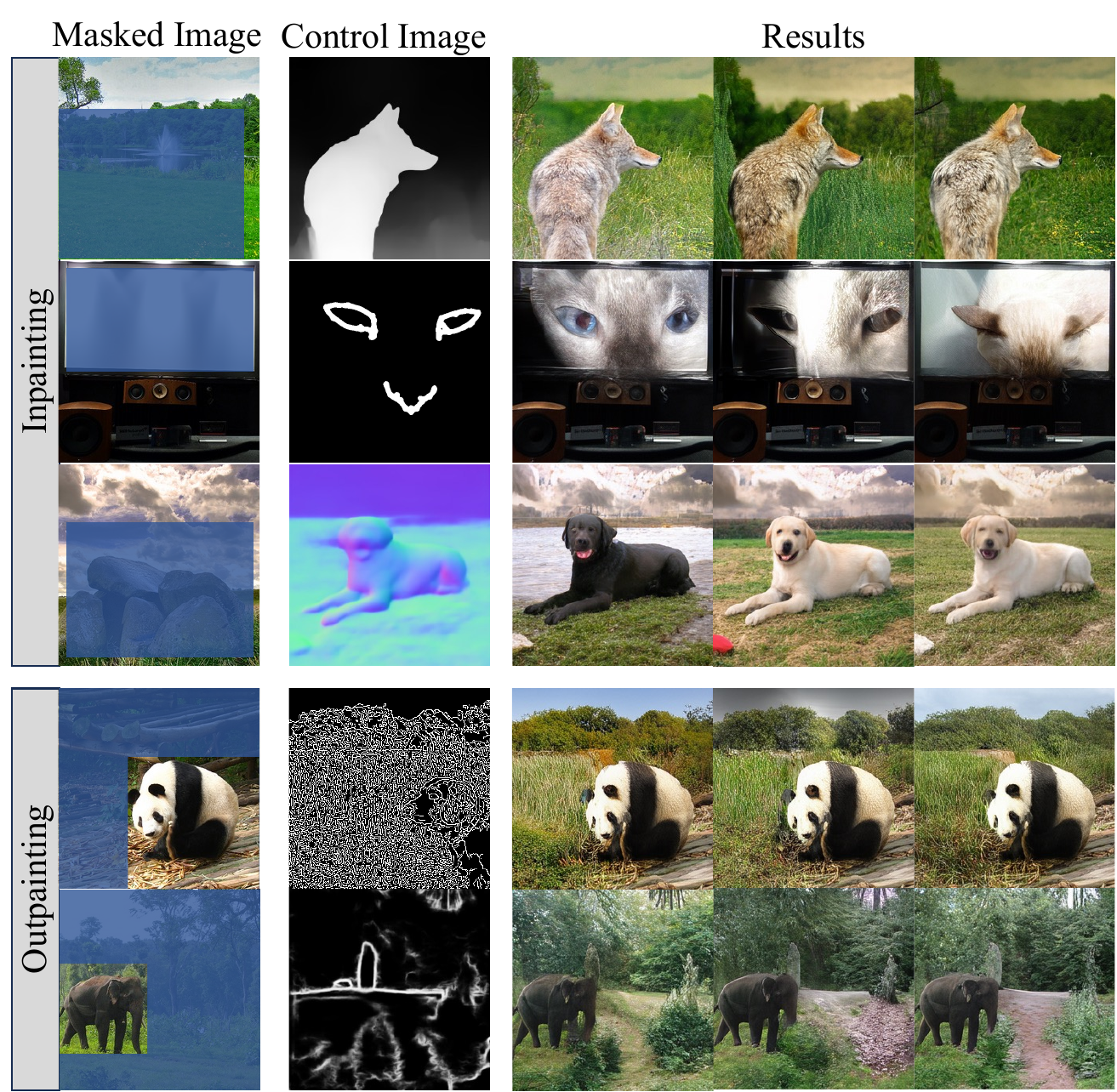}
\caption{
Visualization of zero-shot class-to-image controllable generation task with SCALAR, including inpainting and outpainting.
From top to bottom, Depth, Sketch, Normal, Canny, and HED are used as control image inputs.
}
\label{fig:edit}
\end{figure}

\section{Results}

\subsection{Experimental Setup}
\noindent \textbf{Dataset.} 
Our experiments are conducted on class-to-image (c2i) controllable generation. We conduct experiments on the ImageNet-256~\cite{deng2009imagenet}, using all 50K images from the validation set for evaluation. 
To assess the controllable generation capability of our model, we consider five types of conditions: Canny~\cite{canny1986computational}, Depth~\cite{ranftl2020towards}, Normal~\cite{vasiljevic2019diode}, HED~\cite{xie2015holistically}, and Sketch~\cite{su2021pixel}.

\noindent \textbf{Training Details.} 
(1) We extract four features from the Image encoder $\mathcal{E}$ at depths $d - 1 - k \cdot \lfloor d/4 \rfloor$ ($k{=}0,1,2,3$).
(2) We use DINOv2~\cite{dinov2} as the image encoder for its strong representations and scalability.
(3) For class-to-image (c2i) generation, we use pretrained VARs~\cite{var} with depths of 12, 16, 20, and 24. DINOv2-S is used for d12, and DINOv2-B otherwise, as detailed in~\Cref{tab:main_result}. In this setting, our SCALAR and SCALAR-Uni are trained for 10 epochs with the AdamW optimizer on 8 NVIDIA H20 GPUs.
(4) Following previous works~\cite{zhang2023adding,li2024controlnet++}, the control block is zero-initialized.
(5) For SCALAR-Uni, we sample different control conditions with equal probability.

\noindent \textbf{Evaluation Metrics.}
We mainly employ two metrics: conditional consistency and image generation quality.
We evaluate the conditional consistency by calculating the similarity between the input condition images and the extracted condition images from the generated images.
Specifically, we use F1-Score to assess the similarity of Canny and Sketch, Root Mean Square Error (RMSE) for Normal and Depth, Structural Similarity Index Measure (SSIM) for HED and Lineart, and the mean Intersection-over-Union (mIoU) for Segmentation.
For image generation quality, we use Fréchet Inception Distance (FID)~\cite{heusel2017gans} and Inception Score (IS)~\cite{salimans2016improved}.

\subsection{Class-to-image Experimental Results}
We evaluate the class-to-image controllable generation performance of our proposed SCALAR and SCALAR-Uni on the ImageNet~\cite{deng2009imagenet}. As shown in~\Cref{tab:main_result}, we compute both conditional consistency and image generation quality for the images generated by SCALAR and SCALAR-Uni. Furthermore, we compare our approach with existing controllable generation methods, including diffusion-based models~\cite{zhang2023adding, mou2024t2i}, raster-scan AR models~\cite{li2024controlar}, and VAR-based methods~\cite{controlvar, yao2024car}.

\noindent \textbf{Results for SCALAR.} 
Results show that SCALAR consistently outperforms existing methods in image generation quality, achieving superior FID and IS scores. 
Notably, even SCALAR-d12 achieves better FID compared to ControlAR with LlamaGen-L, while VAR-d12 uses only \textbf{50.1\%} of the parameters of LlamaGen-L (e.g., FID on Canny: \textbf{3.12} vs. 7.85; Depth: \textbf{3.83} vs. 4.19).
In terms of conditional consistency, SCALAR also demonstrates competitive performance against other State-of-the-Art (SoTA) methods. Moreover, as the depth of the underlying VAR increases, both the generation quality and the conditional consistency of SCALAR steadily improve. Qualitative visualizations of the generated results are shown in~\Cref{fig:scalar_vis}. 
Besides, SCALAR also demonstrates strong generalization by synthesizing an object of the target class while adhering to the spatial structure of the source control image, even when the label and control structure are inconsistent, as illustrated in \Cref{fig:cls}.

\noindent \textbf{Zero-shot Controllable In- and Outpainting.} 
SCALAR is tested.
For in-painting and out-painting, we teacher-force ground-truth tokens outside the control-image-guided mask. The model is required to generate tokens only within the masked region. During this process, class label information from the control images is also injected. As shown in~\Cref{fig:edit}, without any retraining, SCALAR achieves strong performance on these downstream tasks, further validating its generalization capability.

\begin{figure}[!t]
\centering
\includegraphics[width=1.0\columnwidth]{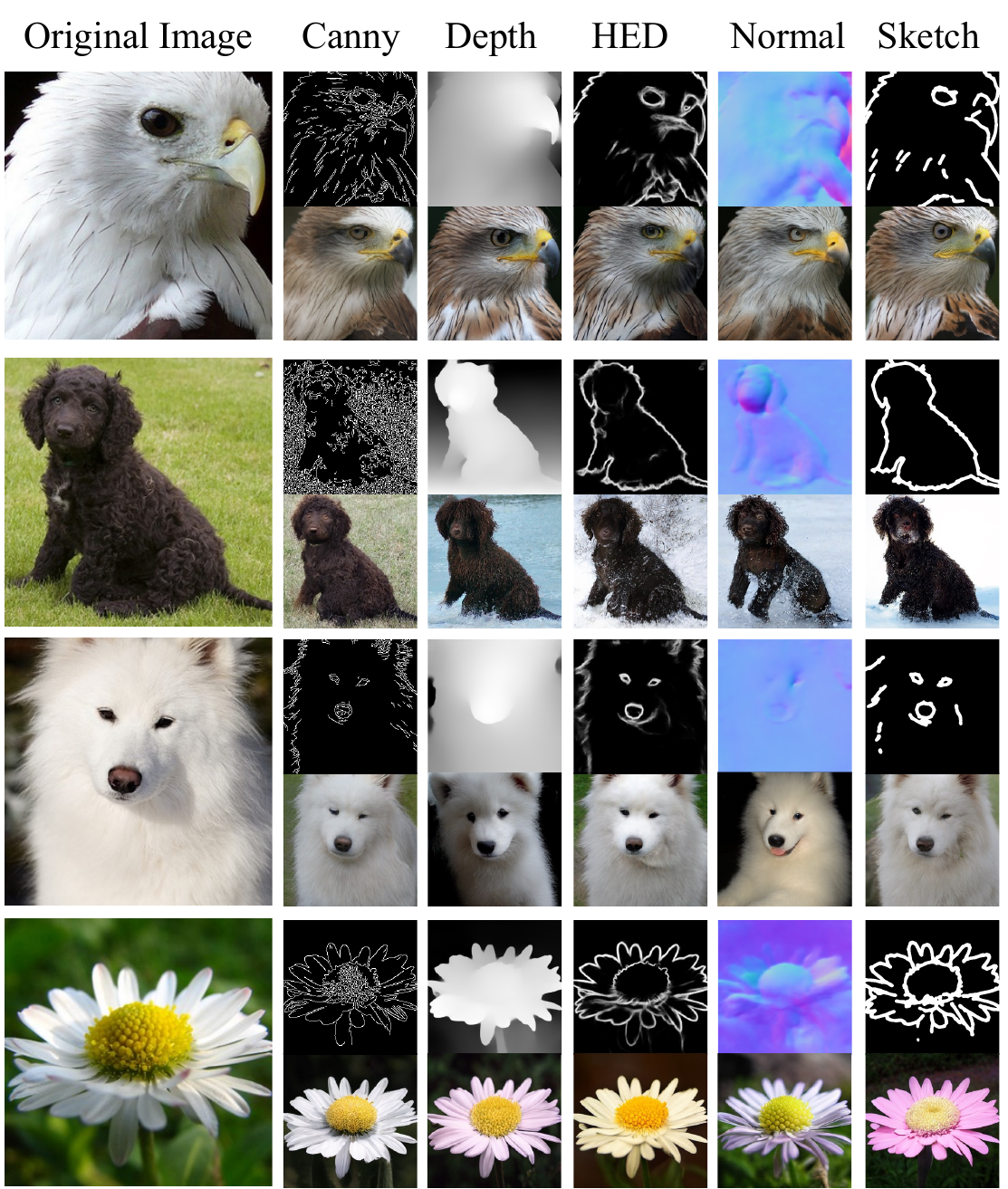}
\vspace{-1.5em}
\caption{
Visual results generated by our unified multi-condition control method SCALAR-Uni under varying control conditions for class-to-image controllable generation.
}
\label{fig:scalar_uni_vis}
\end{figure}
\noindent \textbf{Results for SCALAR-Uni.} 
Compared to other SoTA methods~\cite{zhang2023adding,mou2024t2i,li2024controlar,controlvar,yao2024car}, SCALAR-Uni demonstrates clear advantages in image generation quality while maintaining high conditional consistency. When compared to SCALAR, SCALAR-Uni shows a slight drop in both generation quality and consistency, which may be attributed to the need to accommodate multiple control conditions, thereby reducing the effective training data per condition. Overall, SCALAR-Uni exhibits strong performance and generalization as a unified framework for multi-condition controllable generation. More visualization details can be found in~\Cref{fig:scalar_uni_vis}.

\begin{figure}[!t]
\centering
\includegraphics[width=1.0\columnwidth]{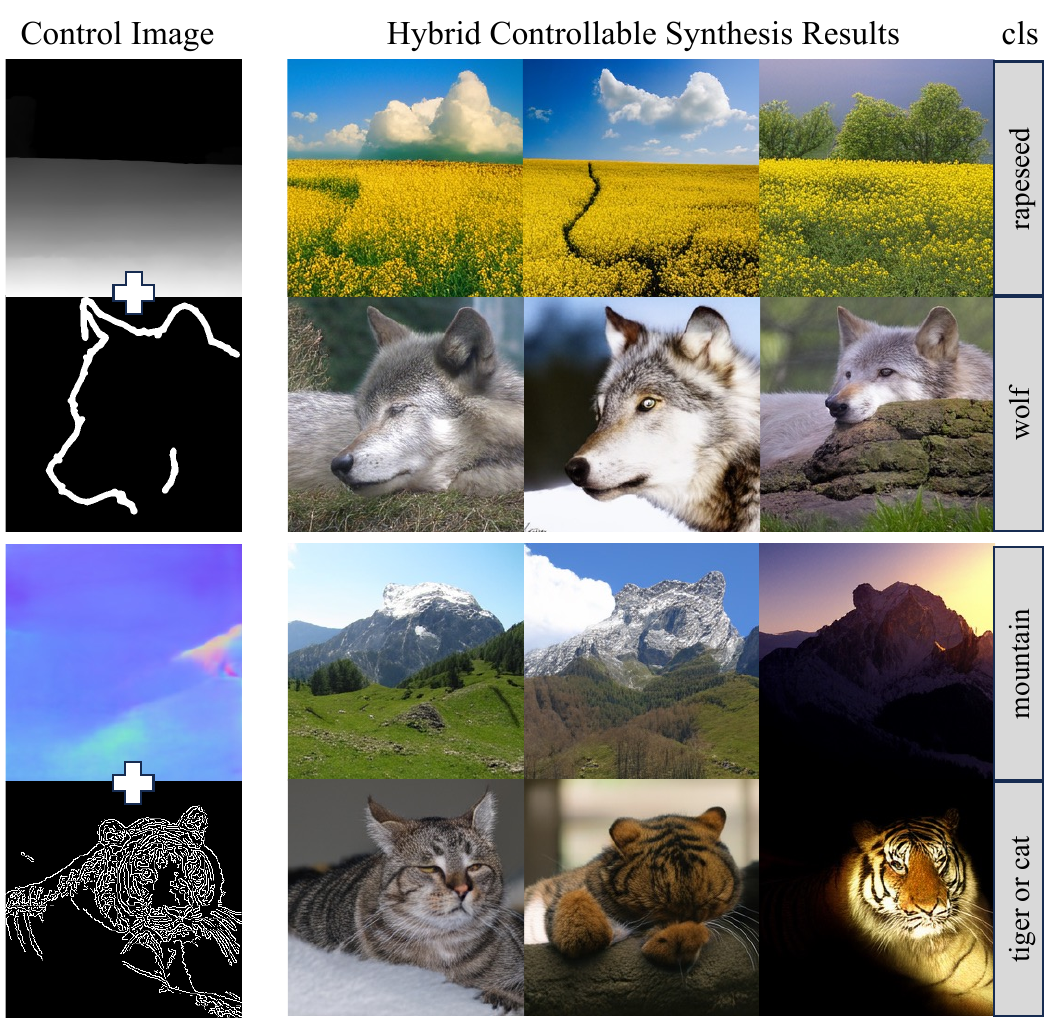}
\caption{
Visualization of zero-shot hybrid class-to-image controllable generation with \textbf{SCALAR-Uni}, combining two control types (e.g., Depth+Sketch) to generate images with both characteristics. 
Different class-condition injections are applied for each control pair.
}
\label{fig:hybrid}
\end{figure}

\noindent \textbf{Zero-shot Hybrid Controllable Synthesis.} 
SCALAR-Uni is tested.
Given the Unified Control Alignment strategy of SCALAR-Uni, we consider two different categories of control images as inputs (e.g., Depth+Sketch, Normal+Canny, etc.). 
In addition to the corresponding control images, the class conditions of both inputs are also fed into SCALAR-Uni. 
As in the zero-shot tasks of SCALAR, no retraining is conducted. 
As shown in~\Cref{fig:hybrid}, SCALAR-Uni successfully combines different control types and generates high-quality images with the characteristics of both controls.
This experiment demonstrates the strong generalization ability of SCALAR-Uni, where the Unified Control Alignment effectively maps diverse control features into a common, modality-agnostic latent space.

\subsection{Ablation Study}
\begin{figure}[!t]
\centering
\includegraphics[width=1.0\columnwidth]{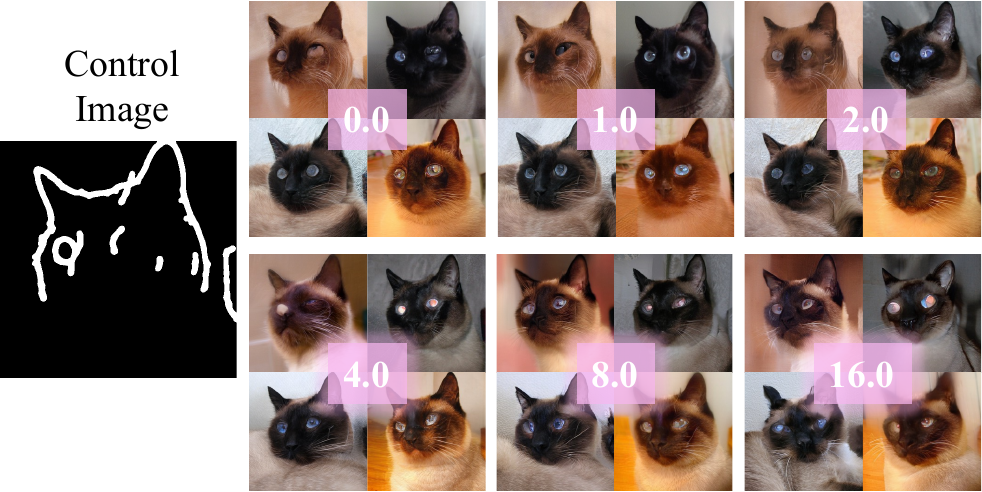}
\caption{
Visualization of images generated with different guidance scales. 
All images at the same position are generated by the same random seed.
}
\label{fig:cfg}
\end{figure}

\noindent \textbf{Classify Free Guidance.}
Given the same control image, we further visualize the results generated with different guidance scales in~\Cref{fig:cfg}. 
We observe that both excessively low and high guidance scales result in lower-quality images. 
Therefore, we set the guidance scale to 4.0 in our experiments.

\noindent \textbf{Ablations on Image Encoder $\mathcal{E}$.}
In~\Cref{tab:image_encoder}, we conduct experiments using different image encoders (or pretraining schemes) towards different controls on ImageNet~\cite{deng2009imagenet}, including Canny and Depth.
Firstly, unlike previous approaches~\cite{li2024controlar,yao2024car}, SCALAR and SCALAR-Uni adopt frozen image encoders~\cite{vit,dinov2,sam} to preserve the robust features obtained from large-scale self-supervised pretraining. 
In addition, inspired by prior work~\cite{ye2024biggait,bolya2025perception,ye2025biggergait}, we extract multi-layer features including intermediate layers, instead of relying solely on the final output, which improves image generation quality while maintaining control consistency.
As shown in~\Cref{tab:image_encoder} (c) to (g), DINOv2~\cite{dinov2} outperforms other pretrained vision models such as ViT~\cite{vit} and SAM~\cite{sam}.
Furthermore, comparing (d), (e), (c), and (g) indicates that increasing the scale of the image encoder significantly boosts performance under the same backbone.
Considering the trade-off between parameter size and effectiveness, we use DINOv2-s for the d12 architecture and DINOv2-b for all other settings.
Finally, the comparison from (g) to (j) demonstrates that deeper VAR networks lead to further improvements in both generation quality and control accuracy, suggesting that deeper models can better exploit visual representations.

\noindent \textbf{Ablations on Unified Control Alignment.}
We evaluate our SCALAR-Uni under different depths (d12, d16, d20, and d24) on ImageNet~\cite{deng2009imagenet}. 
While the model is trained with five types of control conditions, the effectiveness of Unified Control Alignment is validated only on Canny and Depth.
As shown in~\Cref{tab:scalar-uni_ablation}, without the Unified Control Alignment, \textbf{SCALAR-Uni} is equivalent to {\setlength{\fboxsep}{1pt}\colorbox{gray!15}{\textcolor{gray}{SCALAR}}} trained with mixed supervision across the five control types. 
Thanks to the strong general visual representation capability of DINOv2, {\setlength{\fboxsep}{1pt}\colorbox{gray!15}{\textcolor{gray}{SCALAR}}} can effectively capture the differences across modalities and achieve preliminary alignment of multi-modal control signals.
After incorporating Unified Control Alignment, \textbf{SCALAR-Uni} consistently outperforms {\setlength{\fboxsep}{1pt}\colorbox{gray!15}{\textcolor{gray}{SCALAR}}} across different network depths in both image generation quality and control consistency. 
Results in~\Cref{tab:scalar-uni_ablation} demonstrate that \textbf{SCALAR-Uni}, constructed by integrating {\setlength{\fboxsep}{1pt}\colorbox{gray!15}{\textcolor{gray}{SCALAR}}} with Unified Control Alignment, offers a simple and effective solution for controllable image generation under diverse conditions.

\section{Conclusion and Future Work}
\noindent\textbf{Conclusion.} 
In this work, we propose SCALAR, a controllable generation method based on VAR that introduces a Scale-wise Conditional Decoding mechanism adapted for the hierarchical nature of VAR models. By leveraging a pretrained image encoder to extract semantically rich control features and injecting them into scale-specific layers, SCALAR enables persistent and structured guidance throughout the generation process. Our extension, SCALAR-Uni, further supports unified multi-conditional control. We hope our findings will inspire further research on controllable generation within Visual Autoregressive models and encourage the exploration of scale-aware design in future generative methods.

\noindent\textbf{Future Work.} 
(1) To align with the next-scale paradigm in VAR, we directly inject control features extracted from the image encoder into corresponding scales after resizing them via bilinear interpolation. 
While simple to implement, this approach is relatively coarse and lacks fine-grained semantic alignment, which may limit the further improvement of SCALAR's controllability.
This issue is worth further study.
(2) SCALAR successfully combines the powerful generation capabilities of VAR with a simple and effective control injection mechanism. 
We are excited to inspire related tasks (e.g., image editing).

\begin{table}[!t]
\centering
\caption{
Ablation of Image Encoder $\mathcal{E}$ and the backbone depth of VAR.
\colorbox[HTML]{FFFFC8}{\phantom{xx}} cells mark Image Encoder parameters comparable to CAR~\cite{yao2024car} / ControlAR~\cite{li2024controlar}, with a smaller backbone.
}
\renewcommand{\arraystretch}{1.4}
\vspace{-0.4em}
\resizebox{1.0\columnwidth}{!}{
\begin{tabular}{c|c|c|c|c|ccc|ccc}
\toprule
\multicolumn{2}{c|}{\multirow{2}{*}{Index}}     & \multirow{2}{*}{\shortstack[c]{Image\\Encoder $\mathcal{E}$}}& \multirow{2}{*}{\shortstack[c]{Para.\\of $\mathcal{E}$}} & \multirow{2}{*}{Backbone} & \multicolumn{3}{c|}{Canny} & \multicolumn{3}{c}{Depth} \\
\multicolumn{2}{c|}{}                           &                                                              &                                                          &                           & FID↓    & IS↑   & F1-Score↑   & FID↓     & IS↑    & RMSE↓    \\ \midrule
\multirow{3}{*}{\rotatebox{90}{ControlAR}}& \multirow{3}{*}{(a)} & \multirow{3}{*}{\shortstack[c]{ViT-S\\\small(unfrozen)}}     & \multirow{3}{*}{21.8M}                                   & AiM-L                     & 9.66    &  -    &  30.36      & 7.39     &   -    & 35.01   \\
                       &                     &                                                              &                                                          & LlamaGen-B                & 10.64   &  -    &  34.15      & 6.67     &   -    & 32.41   \\
                       &                     &                                                              &                                                          & LlamaGen-L                & 7.69    &  -    &  34.91      & 4.19     &   -    & 31.11   \\ \midrule \midrule
\multirow{2}{*}{\rotatebox{90}{CAR}}& \multirow{2}{*}{(b)} & \multirow{2}{*}{\shortstack[c]{Convs\\\small(unfrozen)}}     & \multirow{2}{*}{21.2M}                                   & VAR-d16 &\midtilde 12.8 & \midtilde 85 & - & \midtilde 10.8 & \midtilde 95  & -         \\
                                    &           &                                                              &                                                          & VAR-d30                   & 8.30    &167.3  &   -         & 6.90     & 178.6  &    -    \\ \midrule \midrule
\multirow{8}{*}{\rotatebox{90}{SCALAR (Ours)}} & \cellcolor[rgb]{1,1,0.784}(c) & \cellcolor[rgb]{1,1,0.784}DINOv2-S               &  \cellcolor[rgb]{1,1,0.784}22.1M      & \cellcolor[rgb]{1,1,0.784} VAR-d12                   & \cellcolor[rgb]{1,1,0.784}3.12      & \cellcolor[rgb]{1,1,0.784}191.3     & \cellcolor[rgb]{1,1,0.784}28.96    & \cellcolor[rgb]{1,1,0.784}3.83     & \cellcolor[rgb]{1,1,0.784}233.3    & \cellcolor[rgb]{1,1,0.784}36.03        \\
                       & (d)                    & ViT-S                                                        &  21.8M                                                   & VAR-d12                   & 4.34    & 189.5 & 24.19       & 5.01     & 230.9  & 42.00   \\
                       & (e)                    & ViT-B                                                        &  86.4M                                                   & VAR-d12                   & 4.23    & 193.3 & 25.98       & 4.83     & 237.6  & 40.77   \\
                       & (f)                    & SAM-B                                                        &  89.6M                                                   & VAR-d12                   & 6.13    & 165.1 & 28.41       & 5.72     & 210.0  & 40.41   \\
                       & (g)                    & DINOv2-B                                                     &  86.6M                                                   & VAR-d12                   & 2.95    & 188.5 & 29.37       & 3.68     & 230.6  & 35.50   \\
                       & (h)                    & DINOv2-B                                                     &  86.6M                                                   & VAR-d16                  & 2.45     & 237.7 & 31.74       & 3.63     & 285.8  & 34.77   \\
                       & (i)                    & DINOv2-B                                                     &  86.6M                                                   & VAR-d20                  & 2.34    & 254.3  & 32.84       & 3.61     & 301.3  & 33.34   \\
                       & (j)                    & DINOv2-B                                                     &  86.6M                                                   & VAR-d24                  & 2.14    & 261.8  & 33.14       & 3.09     & 306.9  & 33.13   \\
\bottomrule
\end{tabular}
}
\label{tab:image_encoder}
\end{table}

\begin{table}[!t]
\centering
\caption{
Ablations on SCALAR-Uni regarding the unified control alignment 
$\mathcal{L}_{align}$.
}
\vspace{-0.4em}
\resizebox{1.0\columnwidth}{!}{
\begin{tabular}{c|ccc|ccc}
\toprule
 \multirow{2}{*}{Model}& \multicolumn{3}{c|}{Canny} & \multicolumn{3}{c}{Depth} \\
                                                                                                                                         & FID↓   & IS↑   & F1-Score↑   & FID↓   & IS↑   & RMSE↓   \\  \midrule
  SCALAR-Uni-d12                                                                                                                         & \textbf{3.31} & \textbf{207.4} & \textbf{27.52} & \textbf{4.21} & \textbf{231.5} & \textbf{38.57} \\
\rowcolor{gray!15} \hspace{1em}\raisebox{0.3em}{\rotatebox{180}{\reflectbox{$\curvearrowright$}}} \textit{w/o $\mathcal{L}_{align}$ in eq.~\ref{eq:loss_align}}& \textcolor{gray}{3.73}   & \textcolor{gray}{198.7} & \textcolor{gray}{26.35}       & \textcolor{gray}{4.40}   & \textcolor{gray}{217.6} & \textcolor{gray}{40.79}   \\  \midrule
  SCALAR-Uni-d16                                                                                                                         & \textbf{2.78} & \textbf{262.4} & \textbf{30.76}   & \textbf{3.73}       & \textbf{294.1}      & \textbf{37.24}        \\
\rowcolor{gray!15} \hspace{1em}\raisebox{0.3em}{\rotatebox{180}{\reflectbox{$\curvearrowright$}}} \textit{w/o $\mathcal{L}_{align}$ in eq.~\ref{eq:loss_align}}& \textcolor{gray}{2.88} &\textcolor{gray}{235.0} & \textcolor{gray}{30.23}& \textcolor{gray}{4.09}& \textcolor{gray}{254.6}& \textcolor{gray}{37.85}        \\  \midrule
  SCALAR-Uni-d20                                                                                                                         & \textbf{2.64} & \textbf{276.9}      &  \textbf{31.68} & \textbf{3.69}       & \textbf{310.1}      & \textbf{37.12}        \\
\rowcolor{gray!15} \hspace{1em}\raisebox{0.3em}{\rotatebox{180}{\reflectbox{$\curvearrowright$}}} \textit{w/o $\mathcal{L}_{align}$ in eq.~\ref{eq:loss_align}}& \textcolor{gray}{2.84} & \textcolor{gray}{247.3}& \textcolor{gray}{31.50}& \textcolor{gray}{3.74} & \textcolor{gray}{272.7}      & \textcolor{gray}{36.88}        \\  
\bottomrule
\end{tabular}
}
\label{tab:scalar-uni_ablation}
\end{table}

{
    \small
    \bibliographystyle{ieeenat_fullname}
    \bibliography{main}
}

\end{document}